\documentclass{article}

\usepackage{PRIMEarxiv}

\usepackage[utf8]{inputenc} 
\usepackage[T1]{fontenc}    
\usepackage{hyperref}       
\usepackage{url}            
\usepackage{booktabs}       
\usepackage{amsfonts}       

\usepackage{nicefrac}       
\usepackage{amsmath}
\usepackage{microtype}      
\usepackage{lipsum}
\usepackage{fancyhdr}       
\usepackage{graphicx}       
\graphicspath{{media/}}     
\DeclareMathOperator{\sign}{sign}
\DeclareMathOperator*{\argmax}{argmax}

\title{The race to robustness: exploiting fragile models for urban camouflage and the imperative for machine learning security
}

\author{
  Harriet Farlow \\
  School of Engineering and Information Technology \\
  University of New South Wales \\
  Canberra\\
   \And
  Matthew Garratt \\
  School of Engineering and Information Technology \\
  University of New South Wales \\
  Canberra\\
  \And
  Gavin Mount \\
  School of Humanities and Social Sciences \\
  University of New South Wales \\
  Canberra\\
  \And
  Tim Lynar \\
  School of Engineering and Information Technology \\
  University of New South Wales \\
  Canberra\\
}

\begin{document}
\maketitle

\begin{abstract}
Adversarial Machine Learning (AML) represents the ability to disrupt Machine Learning (ML) algorithms through a range of methods that broadly exploit the architecture of deep learning optimisation. This paper presents Distributed Adversarial Regions (DAR), a novel method that implements distributed instantiations of computer vision-based AML attack methods that may be used to disguise objects from image recognition in both white and black box settings. We consider the context of object detection models used in urban environments, and benchmark the MobileNetV2, NasNetMobile and DenseNet169 models against a subset of relevant images from the ImageNet dataset. We evaluate optimal parameters (size, number and perturbation method), and compare to state-of-the-art AML techniques that perturb the entire image. We find that DARs can cause a reduction in confidence of 40.4\% on average, but with the benefit of not requiring the entire image, or the focal object, to be perturbed. The DAR method is a deliberately simple approach where the intention is to highlight how an adversary with very little skill could attack models that may already be productionised, and to emphasise the fragility of foundational object detection models. We present this as a contribution to the field of ML security as well as AML. This paper contributes a novel adversarial method, an original comparison between DARs and other AML methods, and frames it in a new context - that of urban camouflage and the necessity for ML security and model robustness.
\end{abstract}

\keywords{artificial intelligence \and machine learning 
\and security}

\section{Introduction}\label{sec:intro}
The field of Adversarial Machine Learning (AML) explores the ability to evade, deceive, hijack or leak information from Machine Learning (ML) systems through methods that broadly exploit the architecture of deep machine learning models. AML is a growing concern as ML use in production systems increases, and has been successfully demonstrated across a broad range of ML applications including computer vision and large language models \cite{alsmadi_adversarial_2022}. There is growing recognition that ML systems should be viewed as an attack surface, as cyber systems are, with appropriate risk-based mitigations and assurance mechanisms based on having benchmarked and accepted tolerance levels for AML threats.

The notion of `hacking' algorithms has existed since the deployment of spam filters and search engine optimisation algorithms in the 1990s \cite{papernot_marauders_2018}. However Szegedy and Goodfellow's work on generating image-based adversarial examples in 2014 precipitated the rise of AML as a discipline \cite{szegedy_intriguing_2014, goodfellow_explaining_2015}. They introduced the Fast Gradient Sign Method (FGSM) to specially craft adversarial perturbations that, when added to an image, disguise the focal object from a classifier \cite{szegedy_intriguing_2014, goodfellow_explaining_2015}. Their proverbial adversarial example presents a panda classified as a gibbon with 99.3\% accuracy by the GoogLeNet model, despite displaying no apparent difference to the human eye \cite{goodfellow_explaining_2015}.  

Computer vision models are already deployed in real world environments, particularly in autonomous driving, facial recognition, scientific research, and manufacturing \cite{khan_machine_2021}. The brittleness of many productionised ML systems were demonstrated through adding obstructions to road signs that successfully deceive autonomous vehicles, causing them to ignore stop signs \cite{eykholt_robust_2018}. Other examples include glasses that evade facial recognition models, and a 3-D printed turtle that convinces an ML classifier it is a rifle regardless of the viewing angle or orientation \cite{athalye_obfuscated_2018, sharif_accessorize_2016}.  AML applications in these arenas therefore constitute a real dilemma as these systems have been proven to be extremely vulnerable to even modest perturbations to the input \cite{li_review_2022}. The extensions of such image-based methods into the `real world' is of growing interest, with techniques like adversarial patches, tarps or glasses demonstrated in computer vision applications as first steps towards creating physical embodiments of these pixel-based attacks \cite{sharif_accessorize_2016, biggio_wild_2018,brown_adversarial_2018, wise_developing_2022}. Urban camouflage may refer to both the military application of camouflage patterns to make soldiers and equipment harder to see in built-up areas, as well as the desire to hide from detection in civilian urban contexts. Automatic detection of civilian and military people and objects in urban environments is increasingly being relegated to machines, where they play a crucial role in tasks such as pedestrian detection, traffic monitoring, surveillance and in autonomous vehicles like cars, trucks and drones \cite{pintor_imagenet-patch_2022}.     

This paper contributes to this body of knowledge by demonstrating a novel attack method based on optimisation algorithms to create Distributed Adversarial Regions (DARs). We conduct an original benchmarking activity by testing different DAR parameters against three models - MobileNetV2, NasNetMobile and DenseNet169 - on a subset of images from the ImageNet dataset that are representative of object detection use cases within urban environments. No other work we are aware of explores the efficacy of distributed regions to disguise a separate target object from detection in black box settings. Other methods typically require the entire image or a large portion of the target object to be perturbed, or are dependent on a white box attack setting. Other methods therefore are also much harder to translate to physical analogies, whereas DARs are constructed to explore this intention.  

This work sheds light on the inherent fragility of these models when subjected to exploitation and emphasises the need for sufficient measures to manage the weaknesses of these models.

\section{Related work}

\subsection{Attack methods}

More than ninety AML attack classes currently exist in AML repositories like the Adversarial Robustness Toolbox (ART) and Cleverhans \cite{noauthor_adversarial_2021, noauthor_cleverhans_2021}. Here we present the five methods that underpin the work in this paper. 

\subsubsection{Universal Adversarial Perturbations}
 
The Universal Adversarial Perturbation (UAP) method adds a perturbation vector to an entire image that alters each pixel, for each image, by a unique value, but is not based on the loss function of the target model \cite{biggio_wild_2018}. The aim is that this vector be small enough to be unobserved by humans, but significant enough to fool the classifier. 

It can be expressed by equation 
(\ref{Eq:UAP}), where $\eta$ is the perturbation vector modelled through an upper bound $\epsilon$ on the $l_{p}$-norm and $\theta$ is the classifier function (in this case a neural network).

The original image is X, the ground-truth label is Y, and the adversarial vector created is X' = X + $\eta$.  

\begin{equation}\label{Eq:UAP}
\theta(X + \eta) \ne \theta(X) \hspace{0.5cm}  \hspace{0.5cm} s.t. \parallel \eta \parallel \, \leq \, \epsilon 
\end{equation} 

This technique is the equivalent of adding random noise with a bounded magnitude. Its success exposes the brittleness of many ML models \cite{moosavi-dezfooli_universal_2017,biggio_wild_2018}. 

\subsubsection{The Fast Gradient Sign Method}

The Fast Gradient Sign Method (FGSM), expressed in equation (\ref{Eq:FGSM}), was first proposed by Szegedy and Goodfellow in 2014 \cite{szegedy_intriguing_2014}. This method calculates the gradient of the model's loss function relative to the input. The attack is based on a one step gradient update along the direction of the gradient's sign at each time stamp, clipped by the epsilon value. We can perform a projected gradient descent on the negative loss function where J is the loss function of $\theta$ with respect to (X,Y).

\begin{equation}\label{Eq:FGSM}
\eta = \epsilon \cdot \sign (\nabla_{X}J(\theta,X, Y))
\end{equation}

Its one-shot nature means it is computationally inexpensive but does not calculate the optimal perturbation vector. Despite its relative simplicity it has been effective against many target models \cite{biggio_wild_2018}.

\subsubsection{Projected Gradient Descent}

Projected Gradient Descent (PGD) takes FGSM a step further and instead of a one-step gradient update, takes an iterative approach, as seen in equation (\ref{Eq:PGD}) \cite{biggio_wild_2018}. It calculates the loss with respect to the true classification over many iterations (t), still constrained by the epsilon value \cite{biggio_evasion_2013}.   

\begin{equation}\label{Eq:PGD}
X^{t + 1} = \prod_{X + \eta} (X^t + \epsilon\cdot \sign(\nabla_X J(\theta, X, Y)))
\end{equation}

This method employs the $L_\infty$ norm because it has access to the compute gradients across the entire image. 

\subsubsection{One Pixel Attack}

The One Pixel Attack (OPA) also uses an iterative approach whereby it identifies a single pixel that, when perturbed, is most likely to cause the object to be to misclassified \cite{su_one_2019}. They convolve the original image $X$ with a single adversarial pixel perturbed to be some RGB colour, and identify the pixel location and colour that maximally perturbs the confidence for that image.     

\subsubsection{Adversarial Patch}

The adversarial patch is an image which, placed near or on the target object, causes the classifier to ignore the item in the scene and report the chosen target of the patch \cite{brown_adversarial_2018}. The patch $\hat{p}$ is obtained using a variant of the Expectation over Transformation (EOT) framework as seen in equation (\ref{Eq:Patch}) \cite{athalye_obfuscated_2018}. The patch is trained to optimize the objective function:

\begin{equation}\label{Eq:Patch}
\hat{p} = \argmax_{p} \mathbb{E}_{x\sim X,t\sim T,l\sim L} [\log \Pr(\hat{y}|A(p, x, l, t)]
\end{equation}

where $T$ is a distribution over transformations of the patch, $L$ is a distribution over locations in the image, and $A$ is a patch application operator \cite{brown_adversarial_2018}.

\begin{equation}
\label{Eq:constraint1} 
\parallel p - p_{orig} 
\parallel_{\infty} \,\le\, \epsilon 
\end{equation}

\section{Methodology}

We consider primarily a use case where the target is a model rather than a human observer, reducing the strict constraint that the perturbations be undetectable to a human observer and instead aiming to camouflage them. Our method utilises an optimisation algorithm to construct the DAR based on a method reminiscent of the One Pixel Attack to identify the top 1, 2, 3 and 4 areas of the image to optimally perturb for each circular DAR of diameter (size) $s$ and number $n$. The DAR surface is constructed using an optimisation of Projected Gradient Descent, normalised box filter smoothing and colour filters. PGD employs the $L_\infty$-norm but constrained to each DAR region in this method \cite{biggio_wild_2018}. The benefit of the technique is that we can identify, target and perturb the smallest possible change that will yield the largest impact to the classification confidence. 

\section{Experiments} 

We choose images across a variety of classes that are applicable for urban camouflage settings, including cars, tanks, ships and animals, from the ImageNet dataset, resulting in over 500 image permutations. For each image we measure the impact of setting the diameter of the DARs to 2, 6, 10, 14 or 18 pixels. We assess the impact of either 1, 2, 3 or 4 regions. We test MobileNetV2, NasNetMobile and DenseNet169 as they are benchmark models for the ImageNet dataset of varying sizes - 3.5, 14, and 23 million parameters \cite{keras_team_keras_2023}.  

\subsection{Comparing DARs with state-of-the-art PGD implementation}

\begin{figure*}
\centering
\includegraphics[scale=0.2]{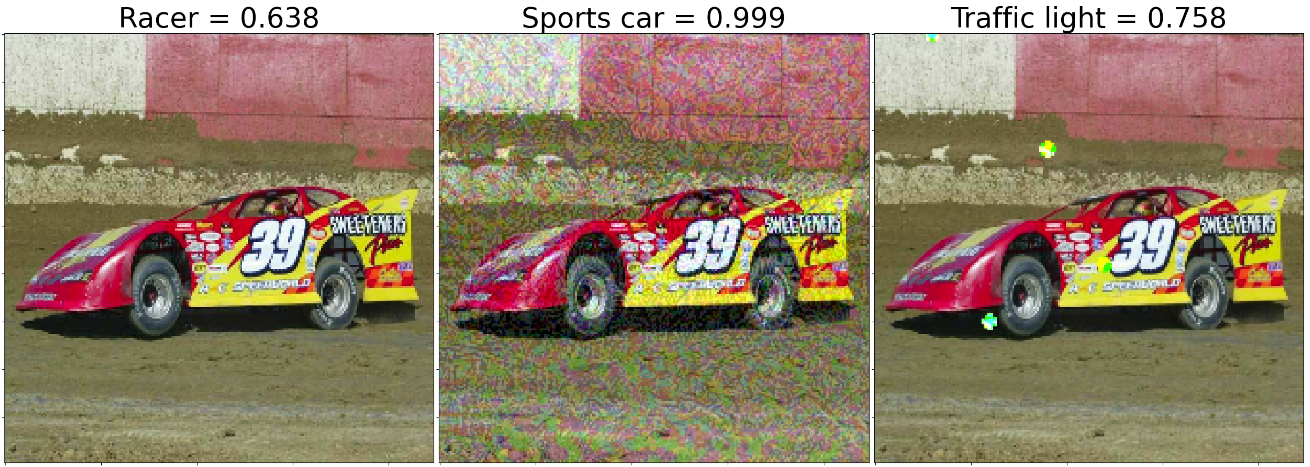}
\caption{Comparing confidence on original image, PGD over whole image, and DARs generated using PGD. This example utilises MobileNetV2 classifier.}
\label{Display}
\end{figure*}

We compare DARs that implement the Universal Adversarial Perturbation (UAP) and Projected Gradient Descent (PGD) methods.  We also compare these results to the state-of-the-art (SOA) implementation of PGD (with epsilon = 2/255), which adds perturbations to the entire image. We see the results in Figures \ref{Graph} and \ref{Display}. With so many different attack types, what is considered SOA does differ between use cases and data type (for example, the Carlini and Wagner method does outperform PGD in white box settings but is less generalisable) \cite{su_is_2018}. For a comprehensive comparison of different SOA attacks we would point the reader to dedicated publications on this topic \cite{su_is_2018}.

As we can see in Figure \ref{Graph}, DARs constructed with PGD are more successful at lowering the top 1 confidence of the three models than UAP, with an average decrease of 40.4\%, although the standard deviation is high at an average of 23.8\%. Attack efficacy is highly dependent on the image, and shows DARs can perform similarly to existing state of the art techniques depending on the circumstance, within 18.1\% on average. However, even DARs constructed using UAP are still successful at lowering the confidence by an average of 21.1\%. In Figure \ref{Display} we show an example of DARs compared to SOA PGD.  

\subsection{Optimising DAR parameters}

We test the impact on classification confidence of holding constant the perturbation method (Projected Gradient Descent), and changing the size of the DARs and the number of DARs applied to each object. As the size of the DAR increases, the classification confidence falls. Likewise, as the number of DARs increases the confidence also decreases. This may seem intuitive, given that adversarial examples typically perturb the entire image, and as the epsilon value increases the confidence similarly decreases. However, validating that this still holds for the small regions represented by DARs is a valuable test. As we can see in Table \ref{Table1}, even one DAR of size 2 decreases the confidence by 19.8\% on average, and four DARs of size 18 decrease the confidence by an average of 58.0\%. 

\subsection{Evaluating DAR generalisability}

\begin{figure*}
\centering
\includegraphics[scale=0.5]{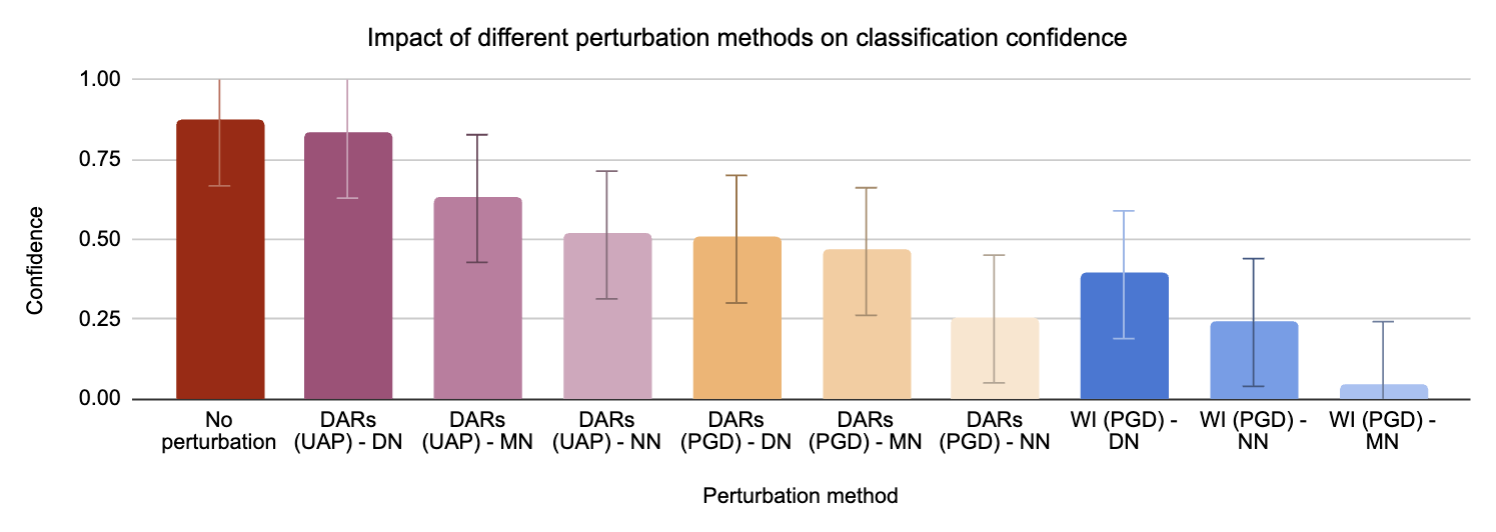}
\caption{Comparing confidence on original image, DARs generated using UAP, DARs generated using PGD, and PGD over whole image against different target models. DN = DenseNet169, MN = MobileNetV2, NN = NasNetMobile.}
\label{Graph}
\end{figure*}

DARs are constructed using gradient magnitudes based on the MobileNetV2 model architecture, therefore targeting MobileNetV2 is a white box attack and validating the DARs against the other two models constitute a black box attack. Figure \ref{Graph} shows that even in black box settings, DARs reduce the confidence in black box settings by 50.0\% against DenseNet169 and NasNetMobile by 25.0\%. 

\begin{table*}[!ht]
    \centering
    \caption{Average confidence for different DAR parameters on ImageNet images}
    \begin{tabular}{|l|l|lll|l|}
    \hline
        \textbf{DARS} & ~ & \textbf{DAR PGD} & ~ & ~ & \textbf{Ave. decrease} \\ \hline
        Size & Number & MobileNetV2 & NasNetMobile & DenseNet169 & ~ \\ \hline
       2 & 1 & 0.515 & 0.711 & 0.782 & 0.198 \\ \hline
        ~ & 2 & 0.520 & 0.694 & 0.803 & 0.195 \\ \hline
        ~ & 3 & 0.432 & 0.571 & 0.771 & 0.276 \\ \hline
        ~ & 4 & 0.435 & 0.339 & 0.737 & 0.363 \\ \hline
        6 & 1 & 0.443 & 0.579 & 0.773 & 0.268 \\ \hline
        ~ & 2 & 0.480 & 0.652 & 0.764 & 0.235 \\ \hline
        ~ & 3 & 0.405 & 0.306 & 0.729 & 0.387 \\ \hline
        ~ & 4 & 0.504 & 0.097 & 0.499 & 0.500 \\ \hline
        10 & 1 & 0.508 & 0.019 & 0.342 & 0.577 \\ \hline
        ~ & 2 & 0.516 & 0.268 & 0.584 & 0.411 \\ \hline
        ~ & 3 & 0.456 & 0.464 & 0.752 & 0.310 \\ \hline
        ~ & 4 & 0.452 & 0.148 & 0.585 & 0.472 \\ \hline
        14 & 1 & 0.466 & 0.034 & 0.288 & 0.604 \\ \hline
        ~ & 2 & 0.444 & 0.000 & 0.104 & 0.684 \\ \hline
        ~ & 3 & 0.431 & 0.162 & 0.432 & 0.525 \\ \hline
        ~ & 4 & 0.421 & 0.405 & 0.740 & 0.345 \\ \hline
        18 & 1 & 0.487 & 0.126 & 0.441 & 0.516 \\ \hline
        ~ & 2 & 0.447 & 0.005 & 0.227 & 0.641 \\ \hline
        ~ & 3 & 0.498 & 0.000 & 0.045 & 0.686 \\ \hline
        ~ & 4 & 0.363 & 0.134 & 0.363 & 0.580 \\ \hline
    \end{tabular}
\label{Table1}
\end{table*}

\section{Findings}

\subsection{DARs do trade off attack efficacy for implementation feasibility, but not significantly}

On average, DARs are less successful than existing SOA PGD perturbations that alter the alter image. However they do, on average, perform within 18.1\%, and this varies widely depending on the image and the target model. As an initial exploration of the DAR method, this indicates that a method as simple as DARs are may be feasible depending on the use case and particularly in circumstances where the entire image cannot be perturbed (for example, as physical objects). 

\subsection{The bigger the perturbation, the more successful the attack, but even a few DARs are successful}
The DAR method is successful at decreasing the likelihood the object will be detected by perturbing as few as one small region within the image frame. As the size of the model grew bigger as measured by the number of parameters (DenseNet169 being the largest at 23 million parameters), it became more robust to perturbations, although was still susceptible. Generally, models that are deployed in the real world will also vary in their robustness to adversarial examples (through their size, architecture, and training process) and may in turn be more or less susceptible to DARs \cite{biggio_wild_2018}. However, there is no standard benchmark that must be met by models before they are deployed that means DARs and other adversarial examples may be successful for many models already in production.

\subsection{Many models are brittle and even random perturbations are often successful}

While PGD was the most successful attack method for DAR construction, UAP was still able to decrease classification confidence by an average of 21.1\%. Since UAP represents small regions of random noise, this highlights how these models are still brittle to `unknowns'. Further, this method highlights the region of the image that is most impactful to the prediction and such information could be useful in improving the model robustness. For example, in the ImageNet images the areas of turbulent water around the ship were found to have a high impact on the classification. This over-reliance on context to the classification of an object is an example of a vulnerability. Therefore, application of DARs could be of value for use in adversarial testing.

\subsection{DARs are highly generalisable to black box settings}

The perturbations were based on MobileNetV2 as a victim model, but were successful at reducing the confidence to almost the same level in the other two models - 46.1\% for MobileNetV2, 25.0\% for NasNetMobile and 50.0\% for DenseNet169. Since NasNetMobile and DenseNet169 represent black box settings, this shows the DARs are generalisable. The impact of DARs is particularly pronounced in situations where classification based on the unperturbed images are already at a low confidence. Where models already struggle to identify an object, application of DARs will further render the object undetectable.

\subsection{Physical manifestations of DAR attacks in this use case are possible and concerning}

The DAR method could be seen from a few different angles: as an attack, a disguise or a benchmarking technique. In the field of AML, many techniques are evaluated as attacks in order to inform threat modelling. In this case, we imagine how an adversary may use a DAR so we can mitigate against this threat. There are many circumstances where an attacker would not want to perturb the object in question, whether for business, security or cost reasons. Static patches can be costly and difficult to remove or alter if algorithms become robust to specific patches under adversarial training. However the placement of DARs by nature are dynamic and their relative position to the object can change, making them more resilient to AML defences.  

\section{Discussion}
The success of the DARs serve as a stark reminder of the fragility and susceptibility of machine learning models, shedding light on the pressing need for robust ML security measures. Despite this simple, almost naive, approach DARs expose the ease with which models can be deceived, particularly in urban camouflage scenarios. They highlight how an adversary with very little skill could affect the kinds of models that may already be productionised. While there certainly exist more complicated AML methods, we propose DARs as one of many possible analogies to the Distributed Denial of Service (DDOS) attack on cyber systems - an attack that requires almost no technical skill but can successfully impact most systems. This vulnerability underscores the urgency to develop effective defences against adversarial attacks.

The transferability of image-based adversarial machine learning techniques to other domains (for example, time series signals) underscores their versatility and potential impact beyond the realm of image classification. Existing adversarial defences encompass a range of techniques, including adversarial training, defensive distillation, input pre-processing, and ensemble methods \cite{noauthor_survey_nodate}. While these defenses offer varying degrees of effectiveness, to bolster ML security, the field needs to pivot from one framed as attacks versus defences, to one of risk-based controls and mitigations, as cyber security does. The establishment of standardised evaluation benchmarks and protocols is vital. These benchmarks can facilitate the objective assessment of existing defences, encouraging the development of more effective techniques \cite{siddiqui_benchmarking_2020}. Policy considerations must be emphasised, and human involvement in the model creation process should be incorporated to enhance the OODA (Observe, Orient, Decide, Act) loop.

Object detection technology is set to play a crucial role in both military and civilian settings. Governments are currently investing in uses that assist military forces in surveillance, reconnaissance, and target acquisition through identification and tracking of enemy vehicles, personnel, and other situational targets \cite{noauthor_attacking_nodate}. Object detection can also assist in the deployment of countermeasures including anti-aircraft systems or perimeter security. In civilian settings, object detection is used in autonomous vehicles, video surveillance and crowd monitoring\cite{noauthor_attacking_nodate}. There are many potential benefits to such technologies, but also introduces risk if these systems are compromised.  

The current landscape is marked by an arms race that is emerging between adversarial machine learning techniques and model robustness. Unfortunately, robustness is often still an afterthought among the ML community \cite{kumar_adversarial_2021}. Governments have released frameworks and regulations focusing on ethical considerations, privacy protection, and risk mitigation in the context of AI safety \cite{noauthor_attacking_nodate}. For instance, the European Union's General Data Protection Regulation (GDPR) outlines principles for data protection relevant to ML systems. Tech companies, such as Google, Microsoft, and OpenAI, claim to utilise ethical frameworks while being in the position of exerting control over its development and access. Concerns arise regarding data access, algorithmic transparency, and concentration of power \cite{noauthor_attacking_nodate}. However ML security, which comprises different risks and mitigations to the safety problem, is still not a widely acknowledged challenge. The release of ChatGPT in November 2022 elevated ML security in the international consciousness and is starting to change that \cite{yampolskiy_artificial_2016}. The emergence of actual attacks on these models, as evidenced by incidents documented in AI incident databases like that by AIAAIC (AI, Algorithmic, and Automation Incidents and Controversies), necessitate urgent action in strengthening ML security.

\section{Future work} 

Future extensions of this work include those things that will help to understand the feasibility of extending this work into real world applications, improving DAR camouflage, and highlighting assurance measures from such methods. These include: creating physical renditions of these objects, exploring the impact of object rotation and distance, benchmarking against other models, and testing a targeted version. 

In particular, it highlights the fragility inherent in many models and urges caution from both a technical and implementation perspective. With ML technologies advancing at an accelerating rate, research into MLOps and processes that assure model robustness must also be an active area of research. The field of ML security must emphasize interdisciplinary collaboration, involving experts from fields such as computer security, adversarial machine learning, evolutionary science, law and the humanities. Otherwise, ML security may follow the lead of cyber security in becoming both a technical, existential and geostrategic threat.     

\section{Conclusion}

In conclusion, this paper proposes a novel instantiation of existing AML methods we refer to as Distributed Adversarial Regions (DARs). The method is benchmarked against three models - MobileNetV2, NasNetMobile and DenseNet169 - on images from the ImageNet dataset in the context of urban camouflage, compared to existing state of the art methods, and evaluated across different DAR hyperparameters (size, number and perturbation method). We find that DARs do still reduce the confidence by an average of 40.4\% without needing to perturb the entire image, and are highly generalisable to black box settings. We present this work as a step towards benchmarking and furthering ML security and assurance.

\section*{Acknowledgments}
I would like to thank Simon Watt for their assistance.

\bibliographystyle{unsrt}  
\bibliography{references}

\end{document}